\documentclass[11pt]{article}

\usepackage[final]{acl}

\usepackage{times}
\usepackage{latexsym}

\usepackage[T1]{fontenc}

\usepackage[utf8]{inputenc}

\usepackage{microtype}

\usepackage{inconsolata}

\usepackage{graphicx}

\usepackage{amssymb}
\usepackage{amsmath}
\usepackage{algorithm}
\usepackage{algorithmicx}
\usepackage{algpseudocode}
\usepackage{booktabs}
\usepackage[dvipsnames]{xcolor}
\usepackage{multirow}
\usepackage[most]{tcolorbox}
\usepackage{geometry}
\usepackage{enumitem}

\title{Does RLVR Extend Reasoning Boundaries? \\ Investigating Capability Expansion in Vision-Language Models}

\author{
Minghe Shen$^{1,2}$, Zhuo Zhi$^{1,2}$, Chonghan Liu$^{3}$, Shuo Xing$^{4}$, Zhengzhong Tu$^{4}$, Che Liu$^{5}$\thanks{Corresponding author.} \\
{\small
$^{1}$University College London \quad
$^{2}$Samsung R\&D Institute UK \quad
$^{3}$University of California, Los Angeles} \\
{\small
$^{4}$Texas A\&M University \quad
$^{5}$Imperial College London} \\
{\small
\texttt{minghe.shen.24@ucl.ac.uk} \quad
\texttt{che.liu21@imperial.ac.uk}
}
}
\begin{document}
\maketitle

\begin{abstract}
Recent studies posit that Reinforcement Learning with Verifiable Rewards (RLVR) primarily amplifies behaviors inherent to the pre-training distribution rather than inducing new capabilities, but these insights are predominantly limited to language-only domains, leaving the dynamics of visual-centric spatial reasoning under-explored. To examine the impact of RLVR on the capability boundaries of Vision-Language Models (VLMs), we introduce \textbf{Ariadne}, a controlled framework based on synthetic maze navigation where the reasoning difficulty is precisely regulated by path length and the number of turns. We demonstrate that applying RLVR extends the spatial reasoning boundary, achieving success on problems where the base policy VLM consistently attains $0\%$ accuracy despite increasing pass@k sampling budgets, indicating that the optimized policy successfully navigates search spaces that were effectively unreachable by the base distribution. Furthermore, despite being trained exclusively on synthetic mazes, we evaluate the model on two real-world navigation benchmarks (MapBench and ReasonMap) in a zero-shot setting. The observed improvements in these out-of-domain tasks suggest genuine spatial reasoning capability expansion rather than mere sampling efficiency.
\end{abstract}

\section{Introduction}

Reinforcement Learning with Verifiable Rewards (RLVR) has emerged as a transformative paradigm for enhancing the reasoning capabilities of Large Language Models (LLMs), enabling breakthroughs in mathematics and coding without reliance on human-annotated CoT data~\cite{deepseekai2025, shao2024deepseekmath, Schulman2017ProximalPO, liu2025beyond, wang2025emergent}.
Yet, as these methods gain prominence, a critical debate has surfaced regarding the underlying mechanisms of these improvements: \textit{Does RLVR facilitate the acquisition of novel reasoning primitives, or does it primarily optimize the sampling efficiency of behaviors already latent within the base policy?}

Recent studies in language-dominant tasks suggest the latter \cite{yue2025does}.
Research shows that on math benchmarks, base models often retain sufficient probability mass over valid trajectories, such that RLVR primarily improves sampling efficiency by amplifying existing behaviors rather than extending reasoning boundaries~\cite{yue2025does, xu2025tinymodelbiglogic}.
We argue that this conclusion depends on the evaluation domain. Tasks such as mathematics and coding, which are well represented in pre-training and prone to contamination, constitute high-support regimes where correct solutions already lie within the base distribution~\cite{matton2024leakage, liang2025swe, mirzadeh2025gsmsymbolic, wu2026reasoning}, making RLVR an efficiency mechanism.
In contrast, visual-centric spatial reasoning tasks often fall into low-support regimes, where prior coverage is minimal and valid solutions are effectively absent. For example, in long-horizon navigation, base models fail to produce valid trajectories even with large pass@$k$ budgets~\cite{feng2025can, xing2025can}. Success in such settings indicates that RLVR induces new reasoning trajectories, reflecting genuine capability expansion within spatial reasoning rather than improved sampling.

To probe this distinction, we introduce \textbf{Ariadne}\footnote{Named after the mythological figure who provided the thread to navigate the Labyrinth, symbolizing our focus on guiding VLMs through structured spatial reasoning.}, a controlled experimental framework based on synthetic maze navigation~\cite{dao2025alphamaze}.
Unlike noisy real-world benchmarks, Ariadne allows us to manually characterize the ``reasoning boundary'', the complexity threshold (path length and turns) where the base model's success rate diminishes to zero in high-complexity configurations.
We train VLMs using Group Relative Policy Optimization (GRPO)~\cite{shao2024deepseekmath} within this setting to investigate whether the optimized policy can successfully navigate search spaces that lack effective support in the base distribution.

Our results provide decisive evidence against the efficiency-centric view in the VLM context.
We demonstrate that RLVR-trained models achieve non-trivial success rates on spatial problems where the base model consistently fails ($0\%$ accuracy) despite extensive sampling.
Furthermore, we show that this learned spatial logic is not an artifact of the synthetic environment; the model exhibits zero-shot transfer improvements on real-world map navigation benchmarks (MapBench and ReasonMap), suggesting the acquisition of robust spatial reasoning primitives.

In summary, our contributions are as follows:
\begin{itemize}
    \item We introduce \textbf{Ariadne}, a controlled framework based on synthetic maze navigation designed to probe the reasoning boundaries of VLMs through verifiable rewards and precise difficulty regulation.
    \item We demonstrate that RLVR extends the spatial reasoning boundary beyond the effective support of the base policy, achieving success on complex instances where the base model fails despite increasing pass@$k$ budgets, with validation via zero-shot transfer to real-world navigation tasks.
\end{itemize}

\section{Related Work}

\paragraph{RLVR and Post-Training in LLMs.}
Reinforcement learning post-training has become the standard paradigm for eliciting complex reasoning in Large Language Models (LLMs). Early approaches utilized PPO with learned reward models~\cite{10.5555/3600270.3602281,10.5555/3666122.3669390}, but recent advances have shifted toward Reinforcement Learning with Verifiable Rewards (RLVR) in domains with deterministic correctness, such as mathematics and code~\cite{lambert2025tulu, hui2024qwen2}. Notable implementations, including OpenAI's o1~\cite{jaech2024openai} and DeepSeek-R1~\cite{deepseekai2025}, employ algorithms like Group Relative Policy Optimization (GRPO) to stabilize training without value networks~\cite{shao2024deepseekmath}. These successes have spurred extensive research into optimizing inference-time compute and self-correction strategies~\cite{zhao2025echo, deng2025unlocking, yang2024qwen2, ying2024internlm}.

\paragraph{RLVR in Vision-Language Models.}
The application of RLVR to the vision-language domain is an emerging frontier. Recent studies have applied RLVR to enhance multimodal reasoning in tasks such as Visual Question Answering (VQA) and chart understanding~\cite{huang2025vision, wang2025vl, wan2025srpo}. However, these evaluations remain predominantly language-centric, often focusing on tasks where visual input serves merely as context for symbolic reasoning rather than requiring executable spatial planning. Consequently, it remains an open question whether RLVR meaningfully improves spatial reasoning capabilities in vision-language settings, such as long-horizon navigation, where pre-trained models frequently fail to generate valid action sequences due to a lack of grounded spatial understanding~\cite{11267249, hou2025do, feng2025can, xing2025can}.

\paragraph{The Efficiency vs. Capability Debate.}
A growing body of theoretical work challenges the source of RLVR's gains. In language-only domains, recent analyses argue that RLVR primarily functions as an efficiency mechanism, optimizing the selection of correct reasoning trajectories that already exist within the pre-trained model's latent support, rather than synthesizing novel capabilities~\cite{ye2025emergence, yue2025does, xu2025tinymodelbiglogic}. Specifically, studies show that on math benchmarks, base models often achieve RLVR-level performance given sufficient sampling budgets, implying strong prior coverage~\cite{liu2025understanding, shah2025rethinking}. This phenomenon is likely reinforced by the fact that such domains are extensively represented in pre-training corpora and may be subject to benchmark contamination or memorization effects, further contributing to their high-support nature~\cite{matton2024leakage, liang2025swe}.

Our work serves as a critical counterpoint to this efficiency hypothesis. We investigate RLVR in a spatial reasoning setting where prior coverage is not merely low but effectively absent. By demonstrating that RLVR enables success in these regimes where the base model fails regardless of sampling budget, we provide empirical evidence that RLVR can indeed extend the reasoning boundary, distinguishing capability expansion from mere sampling efficiency.

\section{Method}

Our methodology is designed to examine the hypothesis that RLVR can extend the reasoning boundaries of VLMs beyond the effective reach of the base policy.
To systematically investigate this capability, we introduce \textbf{Ariadne}, a minimal yet fully controllable post-training framework.
Ariadne enables precise characterization of the reasoning boundary by generating synthetic maze navigation tasks where the ground-truth difficulty can be explicitly regulated, allowing us to identify regimes where the base model consistently fails to produce valid solutions.
\subsection{Ariadne: A Probe for Reasoning Boundaries}

\subsubsection{Verifiable Maze Environment}
\begin{figure}[ht]
    \begin{center}
        \includegraphics[width=\columnwidth]{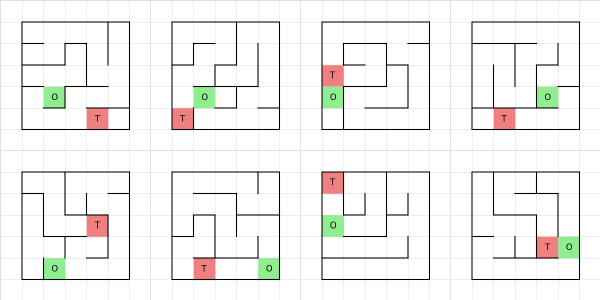}
    \end{center}
    \caption{
    Representative maze instances from Ariadne.
    Green and red cells denote the start and goal, respectively, and black lines indicate walls.
    Mazes vary in path length and number of turns, enabling controlled spatial reasoning difficulty.
    }
    \label{fig:maze_example}
\end{figure}

We construct a controllable testbed based on grid-based maze navigation, adapted from AlphaMaze~\cite{dao2025alphamaze}.
Unlike natural image VQA benchmarks, where the difficulty level or reasoning boundary is hard to explicitly define, Ariadne provides a deterministic environment where the validity of a reasoning step is binary and verifiable.
We generate maze images as visual inputs (see Figure~\ref{fig:maze_example}), creating a pure spatial reasoning task where the VLM is queried to generate the action sequence in textual format (see Appendix~\ref{sec:appendix} for prompt details).
As shown in Figure~\ref{fig:maze_example}, valid solutions require generating a sequence of discrete actions (\texttt{up}, \texttt{down}, \texttt{left}, \texttt{right}) that respect wall constraints.

This design grants us three critical analytical capabilities:
\textbf{(1) Deterministic Verification:} We can explicitly verify every step of a generated chain, eliminating the need for an external judge model that might introduce bias or hallucination.
\textbf{(2) Step-wise Feasibility:} We can distinguish between ``valid but incorrect" paths (wrong turn) and ``hallucinated" paths (walking through walls), allowing for a precise taxonomy of reasoning failures.
\textbf{(3) Parametric Complexity:} By systematically varying path length ($L$) and number of turns ($T$), we can map the \textit{exact} boundary where the base model's pass@$k$ probability collapses to zero.

\subsubsection{Curriculum for Boundary Expansion}
\label{sec:training_dist}
A critical challenge in extending reasoning boundaries is ensuring the model encounters a sufficient density of successful trajectories at the frontier of its capability.
Standard uniform sampling over maze difficulties is inefficient: it allocates excessive capacity to trivial instances where the policy is already competent, while frequently sampling complex instances where the model consistently fails to generate valid solutions.

To bridge the gap between the model's current capability and the target difficulty zone, we design a difficulty-aware curriculum. We control difficulty via optimal path length $s \in \{1, \dots, 5\}$ and sample instances according to an inverted Gaussian-like distribution:
\begin{equation}
    P(s) \propto 1 - \exp\left(-\frac{(s - \mu)^2}{2\sigma^2}\right),
\end{equation}
where $\mu=3$ and $\sigma=2$. This distribution explicitly under-samples the ``comfort zone" (medium difficulty) and over-samples the edges: \textbf{(1) High-Probability Anchors ($s=1,2$)} ensure the model retains basic valid movement primitives; and \textbf{(2) Boundary Frontiers ($s=4,5$)} force the model to attempt tasks at the very edge of its executable horizon.
Crucially, we train exclusively on short-horizon tasks ($L \le 5$, Turns $\le 2$) while evaluating on \textbf{unseen, out-of-distribution (OOD)} instances with significantly extended horizons ($L \le 10$, Turns $\le 4$).
This regime ensures that any observed success reflects genuine generalization to novel complexity levels rather than mere pattern memorization.

\subsection{Policy Optimization via GRPO}

We employ Group Relative Policy Optimization (GRPO)~\cite{shao2024deepseekmath} for VLM post-training.
Formally, for a query $q$, we sample a group of outputs $\{o_1, \dots, o_G\}$ from the reference policy $\pi_{\theta_{\text{old}}}$. The advantage $A_i$ is derived via group normalization of the rewards:
\begin{equation}
    A_i = \frac{r_i - \text{mean}(\{r_1,\dots,r_G\})}{\text{std}(\{r_1,\dots,r_G\})}.
\end{equation}
The policy $\pi_\theta$ is then updated to maximize the surrogate objective:
\begin{equation}
\scalebox{0.9}{$
\begin{aligned}
&\mathcal{J}(\theta) \\
&= \mathbb{E}[ \frac{1}{G} \sum_{i=1}^{G} \min ( \rho_i A_i, \operatorname{clip}(\rho_i, 1-\epsilon, 1+\epsilon) A_i)],
\end{aligned}$}
\end{equation}
where $\rho_i = \frac{\pi_\theta(o_i|q)}{\pi_{\theta_{\text{old}}}(o_i|q)}$ is the importance sampling ratio.
This mechanism incentivizes the model to consistently outperform its average behavior, effectively shifting the probability mass toward higher-reward trajectories.

\subsection{Dense Verifiable Reward Design}

In regimes where the base model lacks support for the correct solution, binary success rewards are insufficient because the model may never reach the goal during early training. To provide a learning gradient, we design a dense, prefix-matching reward function (Algorithm~\ref{alg:correctness}).

Let $A$ be the ground-truth action sequence and $R$ be the predicted sequence. We define the reward $r$ as:
\begin{equation}
\scalebox{0.9}{$
r =
\begin{cases}
\alpha_1 \cdot |A| \cdot \psi(A), & \text{if } R = A \text{ (Success)}\\
\alpha_2 \cdot k \cdot \psi(A_{1:k}), & \text{if } R \neq A \text{ (Partial)},
\end{cases}$}
\end{equation}
where $\alpha_1$ and $\alpha_2$ are weighting factors, $k$ is the length of the longest matching prefix ($A_{1:k} = R_{1:k}$), and $\psi(\cdot)$ is a complexity multiplier based on the number of turns.
This function serves two purposes:
\textbf{(1) Guidance Signal:} It rewards the model for every correct step taken, creating a dense supervision signal that extends the reasoning horizon even if the final goal is missed.
\textbf{(2) Complexity Awareness:} The term $\psi(A)$ scales the reward with structural difficulty (turns), preventing the model from collapsing into simple straight-line heuristics.
This dense signal is the mechanism that allows the policy to climb out of the valley of zero success.

\begin{algorithm}[ht]
\caption{Correctness Reward Function}
\label{alg:correctness}
\begin{algorithmic}[1]
\State rewards $\gets$ []
\ForAll{$(r,a)$ in $(\text{completions}, \text{answers})$}
    \State $r_m \gets$ moves($r$); \quad $a_m \gets$ moves($a$)
    \If{$r_m = a_m$}
        \State reward $\gets |a_m| \times \alpha_1 \times \text{turns}(a_m)$
    \Else
        \State $k \gets$ prefix\_len($r_m, a_m$)
        \State reward $\gets k \times \alpha_2 \times \text{turns}(a_m[1{:}k])$
    \EndIf
    \State rewards.append(reward)
\EndFor
\State \Return rewards
\end{algorithmic}
\end{algorithm}

\section{Experiments}

\subsection{Experimental Setup}

We use Qwen2.5-VL-7B-Instruct~\cite{bai2025qwen25vltechnicalreport} as the primary backbone, and additionally evaluate Qwen2.5-VL-3B-Instruct, Qwen3-VL-4B-Instruct, and Qwen3-VL-8B-Instruct~\cite{bai2025qwen3} to assess generality. Under the Ariadne framework, the model is post-trained with RLVR on 4,700 AlphaMaze samples~\cite{dao2025alphamaze} using the GRPO algorithm.
We additionally train a supervised fine-tuned variant (SFT) on the same data and initialization, with identical training settings but without reinforcement learning.

The reward combines answer correctness (Algorithm~\ref{alg:correctness}), answer format, and reasoning format, with emphasis on correctness ($\alpha_1{=}0.2$, $\alpha_2{=}0.1$).
Training uses 8 NVIDIA A100 (40GB) GPUs with a learning rate of $1\times10^{-6}$ for one epoch, batch size 1, 16-step gradient accumulation (722{,}000 steps), and a warmup ratio of 0.05.

For each training query, we sample a group of $G=8$ candidate responses with a temperature of $1.0$. The GRPO clipping parameter is set to $\epsilon=0.2$. During evaluation across all three benchmarks, each prompt is evaluated using independent roll-outs at a temperature of 1.0, and reported results are averaged across these runs.

\subsection{Benchmarks and Metrics}

We evaluate capability extension across two distinct regimes: (1) internal boundary extension within the controlled Ariadne framework, and (2) zero-shot transfer to external real-world benchmarks.

\begin{figure*}[ht]
\begin{center}
    \includegraphics[width=0.95\textwidth]{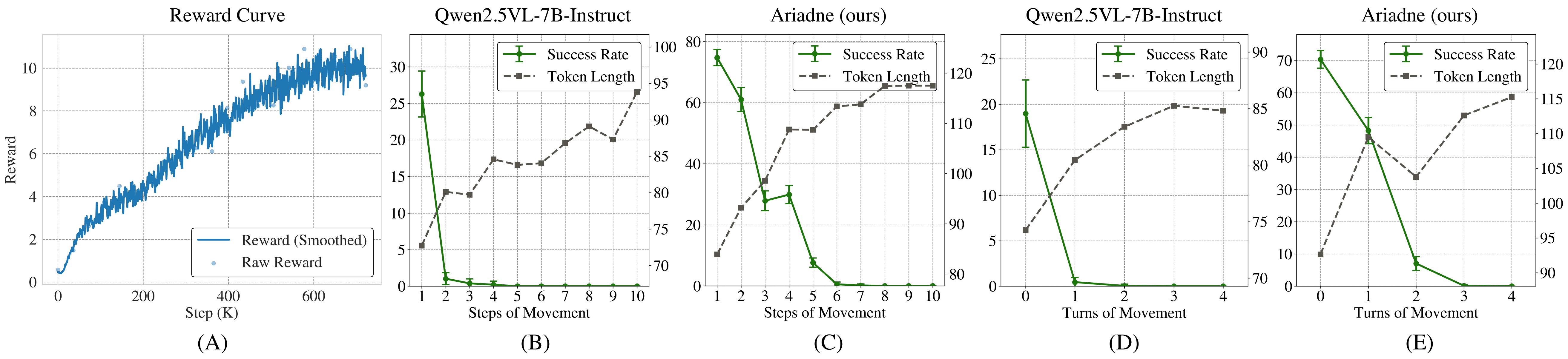}
\end{center}
\caption{
Training dynamics and reasoning boundary shift on AlphaMaze.
\textbf{(A)} GRPO training demonstrates stable reward convergence.
\textbf{(B, D)} The base VLM exhibits a sharp ``capability collapse'' at low complexity (exceeds 2 steps or 1 turn), failing to generate valid paths regardless of token length.
\textbf{(C, E)} Ariadne triples the effective reasoning horizon (shifting the collapse point from 2 to 6 steps) and recovers non-trivial success rates in regimes where the base model has effectively zero support.
}
\label{fig:reward_and_curves}
\end{figure*}

\begin{figure*}[ht]
\begin{center}
    \includegraphics[width=0.95\textwidth]{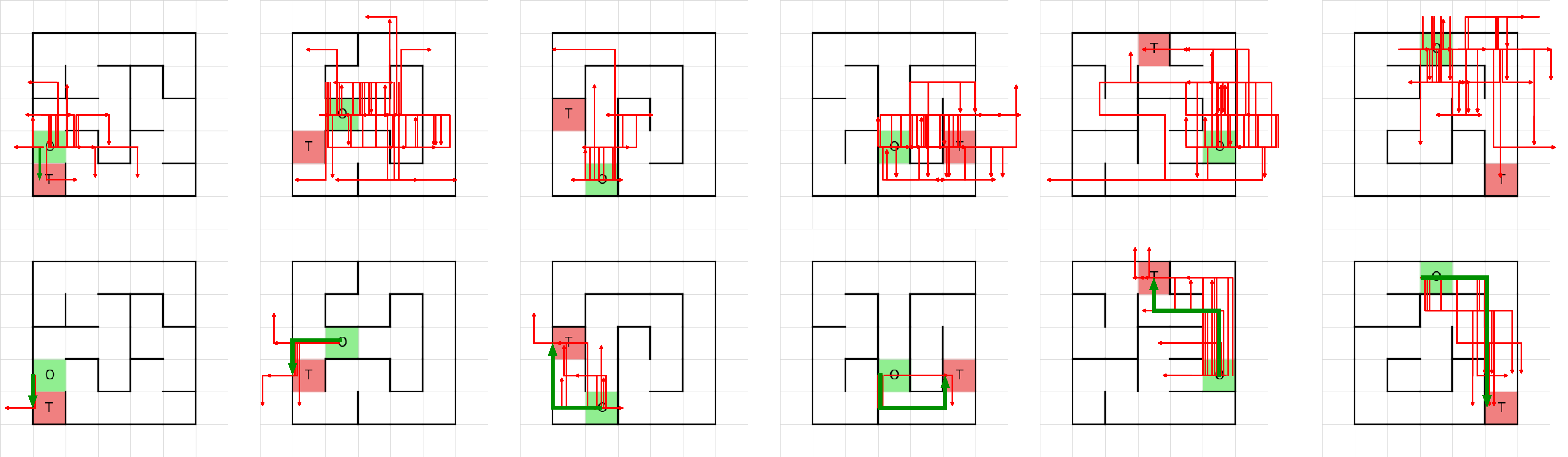}
\end{center}
\caption{
Qualitative comparison of search dynamics ($T=1.0, N=128$).
\textbf{Top:} The base VLM suffers from exploration paralysis, generating diverse but invalid trajectories (\textcolor{red}{red}) that fail to ground in the maze structure.
\textbf{Bottom:} Ariadne produces successful trajectories (\textcolor{OliveGreen}{green}) that the base VLM fails to generate. This visualizes the transition from unstructured failure to structured spatial reasoning.
}
\label{fig:maze_reasoning_paths}
\end{figure*}

\paragraph{Controlled Evaluation (Ariadne).}
To rigorously map the reasoning boundary, we utilize a held-out test set generated via Ariadne (adapted from AlphaMaze~\cite{dao2025alphamaze}). 
Crucially, this set is stratified by difficulty, including "training-distribution" instances (Length $\le 5$) and "out-of-distribution" (OOD) instances (Length $6\text{--}10$, Turns $\le 4$) that were strictly excluded during post-training.
The primary metric is \textbf{Success Rate (SR)}, defined as the percentage of episodes where the generated action sequence exactly matches the ground-truth path.

\paragraph{Real-World Transfer.}
We evaluate generalization on two established benchmarks (see examples in Figure~\ref{fig:mapbench_examples} and~\ref{fig:reasonmap_examples}) that require similar spatial reasoning while differing substantially in visual domains:
\begin{itemize}
    \item \textbf{MapBench}~\cite{xing2025can} evaluates navigation on realistic street maps. We report the \textbf{Shortest Path (SP) Score}, which measures the optimality of the generated path relative to the ground truth (where $1.0$ indicates an optimal path).
    \item \textbf{ReasonMap}~\cite{feng2025can} assesses topological reasoning on complex transit networks. We utilize the weighted \textbf{Map Score}, a metric that aggregates correctness across varying difficulty levels (short vs. long-horizon queries) to reflect robust spatial planning capabilities.
\end{itemize}

\begin{figure*}[ht]
\centering
\includegraphics[width=\textwidth]{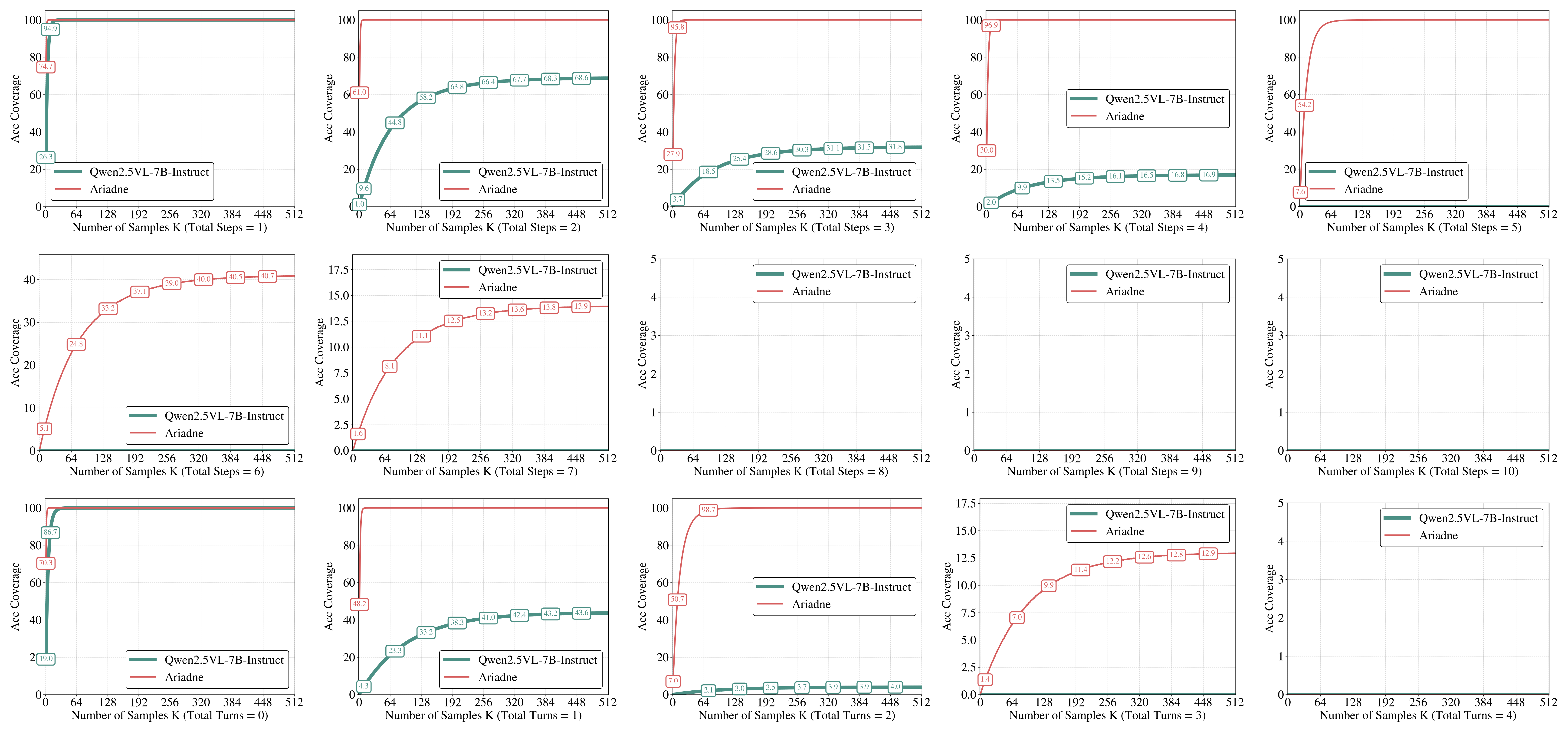}
\caption{Pass@$k$ coverage analysis ($k \le 512$). \textbf{Base Model Saturation:} For complexities $\ge 5$ steps or $\ge 2$ turns, the base VLM saturates near $0\%$, indicating a total absence of valid solutions in its latent space regardless of sampling budget. \textbf{Ariadne Generalization:} Ariadne exhibits logarithmic scaling on unseen complexities (Steps 6--7, Turn 3), confirming robust generalization beyond the training data (Steps $\le 5$, Turns $\le 2$). In the far-OOD regime (Steps $\ge 8$, Turn 4), the performance naturally converges to the intrinsic complexity limit of the learned representations.\protect\footnotemark}
\label{fig:acc_turns_steps}
\end{figure*}

\begin{figure*}[ht]
\begin{center}
\includegraphics[width=0.95\textwidth]{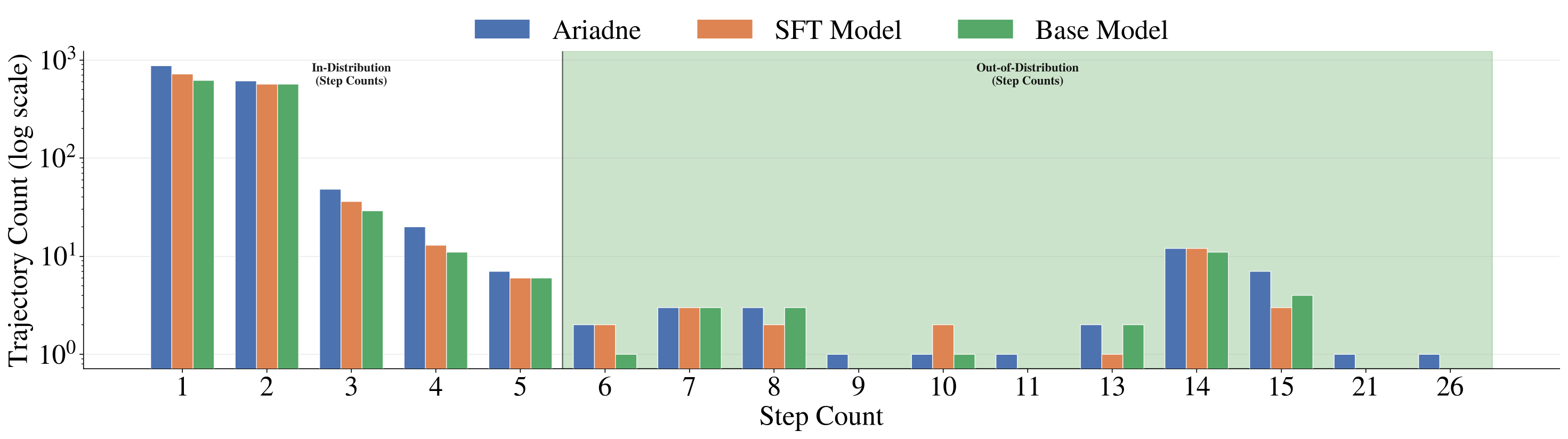}
\end{center}
\caption{Mechanism of transfer on MapBench. \textbf{Left (In-Distribution):} Ariadne significantly amplifies the density of successful trajectories within the 1--5 step range (matching the Ariadne curriculum). \textbf{Right (Out-of-Distribution):} This amplified density ``spills over'' into the OOD regime ($6+$ steps). While the transfer is not a rigid boundary copy, all models achieve some success on longer real-world paths. Ariadne uniquely maintains valid solutions at extreme lengths (e.g., 21, 26 steps) where the base and SFT VLM vanish. Steps with zero success for both models are excluded from visualization.}
\label{fig:mapbench_steps}
\end{figure*}

\subsection{RLVR Extends the Effective Reasoning Boundary}
\label{sec:results_boundary}
We first investigate whether RLVR genuinely expands the model's spatial reasoning capabilities or simply amplifies behaviors already present in the base policy. Figure~\ref{fig:reward_and_curves} illustrates the training dynamics using Qwen2.5-VL-7B-Instruct as the base VLM, along with comparative performance on the AlphaMaze test set.

\paragraph{Stable Optimization and Boundary Shift.}
As shown in Figure~\ref{fig:reward_and_curves} (A), GRPO training yields a steady ascent in reward, indicating stable policy optimization.
The impact is quantified in Panels (B--E): the pretrained base VLM exhibits a distinct \textbf{reasoning boundary}, where the probability of valid navigation collapses to near zero ($<1\%$) once task complexity exceeds 2 steps or 1 turn.
This quantitative collapse is mirrored by the qualitative behavior shown in Figure~\ref{fig:maze_reasoning_paths}. Under identical sampling conditions, the base model (top) generates diverse but unsuccessful trajectories, whereas Ariadne (bottom) produces structured, successful plans.
Post-training, Ariadne fundamentally shifts this boundary, tripling the effective reasoning horizon (from 2 to $\sim6$ steps) and recovering $\sim10\%$ success on 2-turn mazes where the base model consistently fails.

\footnotetext{Plots for Steps $\ge 8$ and Turn 4 appear blank because both the base model and Ariadne achieve a $0\%$ success rate, reflecting the extreme out-of-distribution difficulty.}

\paragraph{RLVR Breaks the Base VLM Reasoning Boundary.}
A critical question regarding RLVR is whether it induces novel reasoning paths or simply improves the sampling efficiency of existing ones.
Figure~\ref{fig:acc_turns_steps} provides decisive evidence against the pure efficiency hypothesis \cite{yue2025does}.
We observe a critical distinction between ``soft'' and ``hard'' complexity regimes.
In the ``soft'' regime (Steps $< 5$), the base VLM's coverage rises with sampling budget, indicating that solutions exist but are rare.
However, this dynamic disappears in the ``hard'' regime:
\begin{itemize}
    \item \textbf{Base Model Collapse:} For tasks beyond the initial boundary (Steps $\ge 5$, Turns $\ge 2$), the base VLM's coverage curve flat-lines at exactly $0\%$ even as $k \to 512$. This ``hard zero'' indicates that valid trajectories are effectively absent from the model's search space, meaning no amount of sampling efficiency optimization could recover the solution.
    \item \textbf{OOD Generalization (Ariadne):} In contrast, Ariadne exhibits robust scaling well outside its training distribution. Despite being trained only on short horizons (Steps $\le 5$, Turns $\le 2$), it surprisingly achieves healthy coverage growth on unseen complexities like Steps 6--7 and Turn 3.
\end{itemize}
This establishes that in ``hard'' regimes, Ariadne is not merely optimizing efficiency, but \textbf{inducing novel behaviors} that bridge the gap between the absence of valid solutions and successful execution.
Finally, we note that this capability extension is finite; in extreme out-of-distribution settings (e.g., Steps $\ge 8$, Turns $\ge 4$), the induced policy approaches its natural complexity horizon, reflecting the fundamental difficulty gap between the training support and far-OOD reasoning.

To further eliminate the possibility that the observed ``hard zero'' phenomenon arises from insufficient sampling, we extend the evaluation budget to pass@1024 in Table~\ref{tab:base_grpo_results}, where the results remain consistent with the trends in Figure~\ref{fig:acc_turns_steps}.

\begin{table*}[!htp]
\centering
\caption{Pass@$k$ coverage across step complexity under extended sampling budgets ($k \le 1024$). Rows denote step counts, and columns report coverage for the base model (Qwen2.5-VL-7B-Instruct) and its Ariadne (RLVR-trained variant) at increasing $k$, including pass@512 and pass@1024.}
\label{tab:base_grpo_results}
\setlength{\tabcolsep}{4pt}
\renewcommand{\arraystretch}{1.15}
\resizebox{0.95\textwidth}{!}{%
\begin{tabular}{c|cc|cc|cc|cc|cc|cc|cc|cc}
\toprule
\multirow{2}{*}{\textbf{N}} 
& \multicolumn{2}{c|}{pass@1}
& \multicolumn{2}{c|}{pass@4}
& \multicolumn{2}{c|}{pass@8}
& \multicolumn{2}{c|}{pass@16}
& \multicolumn{2}{c|}{pass@64}
& \multicolumn{2}{c|}{pass@128}
& \multicolumn{2}{c|}{pass@512}
& \multicolumn{2}{c}{pass@1024} \\
\cline{2-17}
& Base & Ariadne
& Base & Ariadne
& Base & Ariadne
& Base & Ariadne
& Base & Ariadne
& Base & Ariadne
& Base & Ariadne
& Base & Ariadne \\
\midrule
1  & 26.3 & 74.8 & 70.2 & 99.6 & 90.9 & 100.0 & 99.1 & 100.0 & 100.0 & 100.0 & 100.0 & 100.0 & 100.0 & 100.0 & 100.0 & 100.0 \\
2  & 1.1  & 61.1 & 4.0  & 97.5 & 7.8  & 99.9  & 14.5 & 100.0 & 41.2  & 100.0 & 56.8  & 100.0 & 68.8  & 100.0 & 69.0  & 100.0 \\
3  & 0.4  & 27.9 & 1.5  & 72.6 & 3.0  & 92.3  & 5.7  & 99.3  & 16.9  & 100.0 & 24.7  & 100.0 & 31.9  & 100.0 & 32.0  & 100.0 \\
4  & 0.2  & 29.9 & 0.8  & 75.6 & 1.6  & 93.8  & 3.1  & 99.6  & 9.0   & 100.0 & 13.1  & 100.0 & 16.9  & 100.0 & 17.0  & 100.0 \\
5  & 0.0  & 7.6  & 0.0  & 26.9 & 0.0  & 46.5  & 0.0  & 70.9  & 0.0   & 98.9  & 0.0   & 100.0 & 0.0   & 100.0 & 0.0   & 100.0 \\
6  & 0.0  & 0.5  & 0.0  & 2.1  & 0.0  & 4.1   & 0.0  & 7.8   & 0.0   & 22.7  & 0.0   & 32.3  & 0.0   & 40.8  & 0.0   & 41.0 \\
7  & 0.0  & 0.2  & 0.0  & 0.7  & 0.0  & 1.3   & 0.0  & 2.4   & 0.0   & 7.4   & 0.0   & 10.7  & 0.0   & 13.9  & 0.0   & 14.0 \\
8  & 0.0  & 0.0  & 0.0  & 0.0  & 0.0  & 0.0   & 0.0  & 0.0   & 0.0   & 0.0   & 0.0   & 0.0   & 0.0   & 0.0   & 0.0   & 0.0 \\
9  & 0.0  & 0.0  & 0.0  & 0.0  & 0.0  & 0.0   & 0.0  & 0.0   & 0.0   & 0.0   & 0.0   & 0.0   & 0.0   & 0.0   & 0.0   & 0.0 \\
10 & 0.0  & 0.0  & 0.0  & 0.0  & 0.0  & 0.0   & 0.0  & 0.0   & 0.0   & 0.0   & 0.0   & 0.0   & 0.0   & 0.0   & 0.0   & 0.0 \\
\bottomrule
\end{tabular}%
}
\end{table*}

\subsection{Zero-Shot Transfer to Real-World Tasks}
\label{sec:real_world}

We next examine whether the reasoning behaviors learned in the controlled Ariadne setting transfer to real-world navigation and spatial reasoning tasks.
We evaluate three VLMs in a strictly zero-shot setting based on Qwen2.5-VL-7B-Instruct: the pretrained Base Model, a supervised fine-tuned variant (SFT Model) trained on the same AlphaMaze data, and the RLVR-trained model (Ariadne).
Neither SFT nor Ariadne is exposed to MapBench or ReasonMap during training.

\begin{table}[!htp]
\caption{Zero-shot performance comparison on MapBench (Score $\downarrow$, lower is better). \textbf{Bold} indicates the best performance.}
\label{tab: MapBench}
\centering
\resizebox{\linewidth}{!}{
\begin{tabular}{lccc}
\toprule
Metric & Base VLM & SFT VLM & Ariadne \\
\midrule
Google Map $\downarrow$      & 1.61 & 1.80 & \textbf{1.30} \\
Mall $\downarrow$            & 1.43 & 1.44 & \textbf{1.41} \\
Museum $\downarrow$          & 1.43 & 1.39 & \textbf{1.33} \\
National Park $\downarrow$   & 1.86 & 1.85 & \textbf{1.48} \\
Theme Park $\downarrow$      & 1.78 & 1.66 & \textbf{1.46} \\
Trail $\downarrow$           & 2.29 & 1.90 & \textbf{1.47} \\
Campus $\downarrow$          & 1.62 & 1.55 & \textbf{1.29} \\
Urban $\downarrow$           & 1.93 & 1.94 & \textbf{1.91} \\
Zoo $\downarrow$             & 1.68 & 1.64 & \textbf{1.32} \\
\bottomrule
\end{tabular}}
\end{table}

\paragraph{Transfer of Route Optimality (MapBench).}
MapBench evaluates navigation efficiency via the \textit{Shortest Path (SP) Score} (lower is better, $1.0=$ optimal). As detailed in Table~\ref{tab: MapBench}, Ariadne achieves broad improvements over the base and SFT model.
While SFT yields modest improvements in some structured settings (e.g., Campus and Zoo), it fails to match Ariadne's gains in more challenging or unstructured environments. Notably, the largest improvements from Ariadne occur in terrains such as ``Trail'' and ``National Park'', which exhibit sparse landmarks and irregular connectivity. These environments more closely resemble the abstract connectivity patterns of maze navigation than grid-like ``Urban'' layouts, where all models perform similarly. This confirms that RLVR optimized the policy for generic search strategies rather than specific visual pattern matching.

To analyze the mechanism behind real-world transfer, Figure~\ref{fig:mapbench_steps} decomposes success distributions by path length.
Within the \textbf{1--5 step in-distribution range} (Left), Ariadne substantially outperforms both the base model and SFT, closely aligning with the curriculum support of maze-based training.
Beyond this range, we observe a partial generalization into the \textbf{6+ step out-of-distribution regime} (Right), indicating a spillover of learned behaviors.

Unlike the sharp reasoning boundary observed in the synthetic maze domain, real-world navigation exhibits a softer boundary: all models achieve sporadic success on longer paths ($>10$ steps).
Nevertheless, Ariadne consistently dominates the long-tail region, uniquely producing valid trajectories for extreme cases (e.g., 21 and 26 steps) where the base and SFT VLM fail entirely.
RLVR preserves the VLM’s \emph{capacity} for coherent long-horizon reasoning in more complex settings. In contrast, SFT yields only limited improvements in this regime, suggesting that supervised fine-tuning alone is insufficient to induce robust search and path-planning behavior.
Together, these results indicate that RLVR promotes more effective exploration and long-horizon coordination in spatial reasoning tasks beyond what can be acquired through imitation.

\begin{figure*}[!t]
    \begin{center}
        \includegraphics[width=\textwidth]{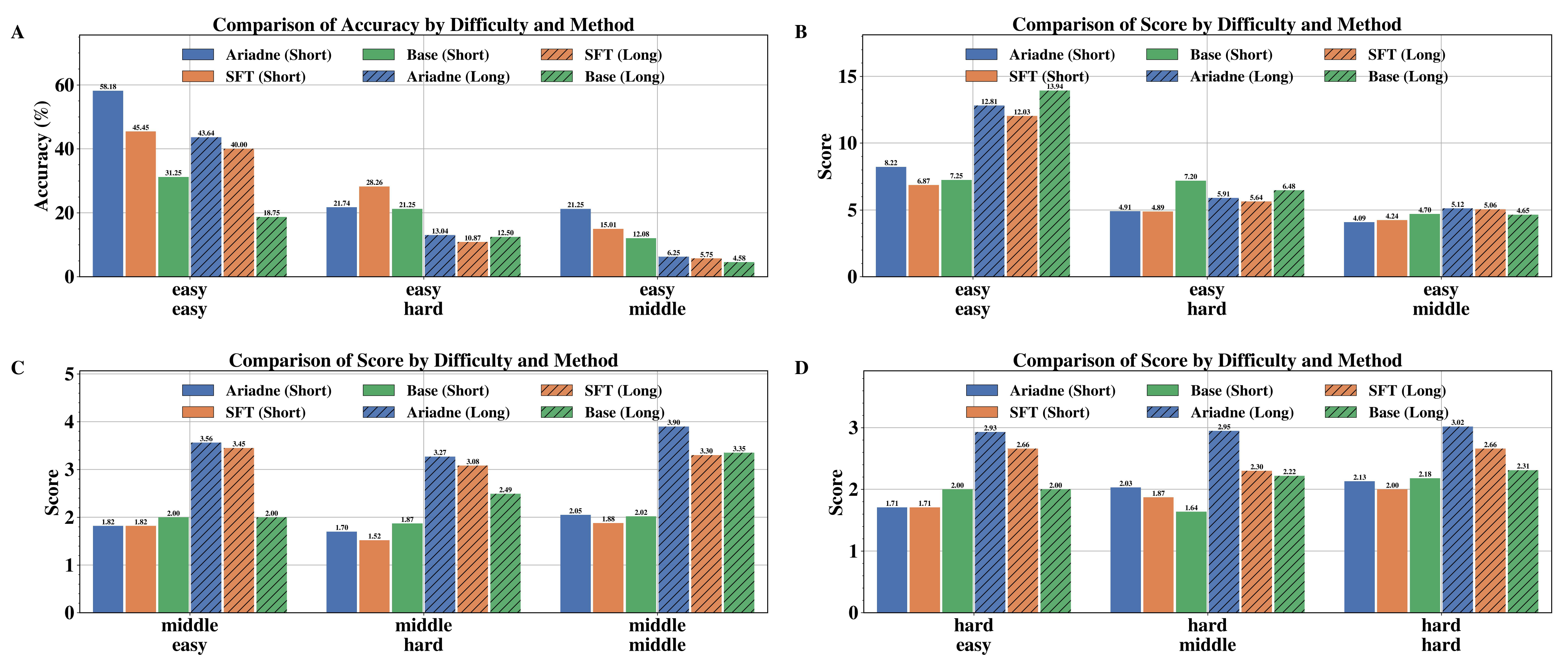}
    \end{center}
    \caption{
    Performance consistency on ReasonMap.
    \textbf{(A)} Ariadne maintains a persistent accuracy lead over the base and SFT VLM.
    \textbf{(B--D)} Map Scores under increasing difficulty. Note that while absolute performance naturally degrades with complexity, Ariadne's advantage is robust, particularly in the long questions in the ``Medium'' difficulty band, mirroring the effectiveness of the difficulty-aware curriculum.
    }
    \label{fig:reasonmap}
\end{figure*}

\paragraph{Scaling Across Model Sizes on MapBench.}

To verify that the observed gains generalize beyond the 7B backbone, we evaluate additional model families and scales, including Qwen2.5-VL-3B-Instruct, Qwen3-VL-4B-Instruct, and Qwen3-VL-8B-Instruct on MapBench. As shown in Table~\ref{tab:mapbench_scale}, Ariadne consistently improves route optimality across all environments and model sizes.

\begin{table}[!htp]
\centering
\caption{Zero-shot performance on MapBench across model scales (Score $\downarrow$). All models are instruction-tuned variants. Lower is better. \textbf{Bold} indicates the better result for each pair.}
\label{tab:mapbench_scale}
\resizebox{\linewidth}{!}{
\begin{tabular}{l|cc|cc|cc}
\toprule
\multirow{2}{*}{Metric}
& \multicolumn{2}{c|}{Qwen2.5-VL-3B} 
& \multicolumn{2}{c|}{Qwen3-VL-4B} 
& \multicolumn{2}{c}{Qwen3-VL-8B} \\
\cline{2-7}
& Base & Ariadne & Base & Ariadne & Base & Ariadne \\
\midrule
Google Map      & 2.70 & \textbf{2.11} & 2.33 & \textbf{2.04} & 1.32 & \textbf{1.25} \\
Mall            & 2.99 & \textbf{2.21} & 2.01 & \textbf{1.65} & 1.22 & \textbf{1.01} \\
Museum          & 2.63 & \textbf{2.37} & 1.93 & \textbf{1.72} & 1.24 & \textbf{1.13} \\
National Park   & 2.24 & \textbf{2.19} & 2.34 & \textbf{1.96} & 1.58 & \textbf{1.29} \\
Theme Park      & 2.61 & \textbf{2.24} & 2.11 & \textbf{1.94} & 1.42 & \textbf{1.16} \\
Trail           & 2.58 & \textbf{2.38} & 2.45 & \textbf{2.32} & 2.08 & \textbf{1.52} \\
Campus          & 2.57 & \textbf{2.07} & 2.13 & \textbf{1.86} & 1.64 & \textbf{1.37} \\
Urban           & 3.12 & \textbf{2.89} & 2.39 & \textbf{2.01} & 1.98 & \textbf{1.79} \\
Zoo             & 2.77 & \textbf{2.09} & 2.52 & \textbf{2.19} & 1.77 & \textbf{1.39} \\
\bottomrule
\end{tabular}}
\end{table}

The improvements are particularly pronounced for smaller models (e.g., 3B and 4B), where the base models exhibit weaker navigation performance. This suggests that RLVR is especially effective in low-capacity regimes, where it compensates for the lack of implicit search structure. As model scale increases, the base models become stronger, but Ariadne continues to provide consistent gains across nearly all environments. Notably, the largest improvements persist in structurally complex settings such as ``Trail'' and ``National Park'', reinforcing that RLVR promotes generalizable search strategies rather than environment-specific heuristics.

\paragraph{Transfer of Reasoning Depth (ReasonMap).}
ReasonMap assesses high-level planning on schematic transit maps. 
Using Qwen2.5-VL-7B-Instruct as the backbone, Table~\ref{tab: ReasonMap} highlights a critical behavioral shift under RLVR: Ariadne not only improves accuracy for long questions but also substantially increases the volume of explicit reasoning.
For Long Questions ($L$), the average token count doubles from $61$ to $121$, accompanied by a $1.47\%$ absolute gain in weighted accuracy and a significant improvement in Map Score ($4.51 \to 5.15$).
In contrast, SFT fails to produce similar gains in the long-question regime, yielding shorter responses and degraded performance.
These results suggest that RLVR promotes explicit, step-by-step reasoning required for complex long-horizon planning, rather than merely increasing selection accuracy.

\begin{table}[!htp]
\caption{Performance comparison on ReasonMap. $S$ indicates short questions, $L$ indicates long questions. \textbf{Bold} indicates the best performance.}
\label{tab: ReasonMap}
\centering
\resizebox{\linewidth}{!}{
\begin{tabular}{lccc}
\toprule
Metric & Base VLM & SFT VLM & Ariadne\\
\midrule
Weighted Acc. ($S$) $\uparrow$      & 13.32$\%$ & \textbf{15.44}$\%$ & 14.50$\%$ \\
$\#$Tokens ($S$)                    & 26 & 25 & 43 \\
Weighted Acc. ($L$) $\uparrow$      & 6.00$\%$ & 4.10$\%$ & \textbf{7.47}$\%$ \\
$\#$Tokens ($L$)                    & 61 & 50 & 121 \\
Weighted Map Score ($S$) $\uparrow$ & 3.73 & \textbf{3.79} & 3.67 \\
Weighted Map Score ($L$) $\uparrow$ & 4.51 & 3.71 & \textbf{5.15} \\
\bottomrule
\end{tabular}}
\end{table}

Figure~\ref{fig:reasonmap} provides a granular breakdown of this performance scaling.
As shown in Panel (a), Ariadne exhibits its largest gains on simpler instances (``Easy/Easy''), where it substantially outperforms both the base model and SFT for Short questions.
Crucially, this advantage persists as task complexity increases.
While absolute performance naturally declines with rising map difficulty (Panels B--D), Ariadne consistently maintains higher Map Scores than the base model across all difficulty tiers.
Notably, even in the hard setting (Panel A, ``Easy/Hard''), Ariadne retains meaningful accuracy ($\sim21\%$), whereas both the base model ($\sim12\%$) and SFT degrade more sharply.
These results indicate that the spatial planning behaviors learned through RLVR transfer to real-world settings and support more robust long-horizon reasoning under increased difficulty.

\paragraph{Scaling of Reasoning Behaviors on ReasonMap.}

We further examine whether the reasoning improvements induced by RLVR persist across model scales beyond the 7B backbone. Table~\ref{tab:reasonmap_scale} shows consistent gains in both accuracy and reasoning depth across Qwen2.5-VL-3B-Instruct, Qwen3-VL-4B-Instruct, and Qwen3-VL-8B-Instruct.

\begin{table}[!htp]
\centering
\caption{Performance on ReasonMap across model scales. All models are instruction-tuned variants. $S$: short questions, $L$: long questions. \textbf{Bold} indicates the better result.}
\label{tab:reasonmap_scale}
\resizebox{\linewidth}{!}{
\begin{tabular}{l|cc|cc|cc}
\toprule
\multirow{2}{*}{Metric} 
& \multicolumn{2}{c|}{Qwen2.5-VL-3B} 
& \multicolumn{2}{c|}{Qwen3-VL-4B} 
& \multicolumn{2}{c}{Qwen3-VL-8B} \\
\cline{2-7}
& Base & Ariadne & Base & Ariadne & Base & Ariadne \\
\midrule
Weighted Acc. ($S$) $\uparrow$ & 7.92$\%$ & \textbf{8.54}$\%$ & 8.05$\%$ & \textbf{8.68}$\%$ & 13.88$\%$ & \textbf{14.96}$\%$ \\
\#Tokens ($S$)                 & 31   & 52            & 24   & 38            & 29    & 47            \\
Weighted Acc. ($L$) $\uparrow$ & 3.48$\%$ & \textbf{4.32}$\%$ & 3.62$\%$ & \textbf{4.54}$\%$ & 6.24$\%$  & \textbf{7.73}$\%$ \\
\#Tokens ($L$)                 & 69   & 134           & 58   & 109           & 64    & 128           \\
Weighted Map Score ($S$) $\uparrow$ & 2.19 & \textbf{2.27} & 2.24 & \textbf{2.31} & 3.82 & \textbf{3.89} \\
Weighted Map Score ($L$) $\uparrow$ & 2.74 & \textbf{3.11} & 2.81 & \textbf{3.20} & 4.68 & \textbf{5.33} \\
\bottomrule
\end{tabular}}
\end{table}

Across all models, Ariadne achieves consistent improvements, particularly in the long-question regime where planning depth is critical. Performance gains are most pronounced in this setting, and while larger models achieve higher absolute performance, the relative improvements from RLVR remain stable, suggesting that the induced behaviors are not tied to a specific capacity regime but generalize across model scales.

\section{Conclusion}

In this work, we investigate the prevailing view that RLVR merely amplifies existing behaviors, providing decisive evidence that it functions as a mechanism for \textbf{inducing novel behaviors} in vision-language domains. By isolating the effective reasoning boundary, where the base model's valid solution space is effectively empty regardless of sampling budget, we demonstrate that RLVR constructs novel spatial primitives that enable success in regimes characterized by the absence of valid solutions. Crucially, these learned behaviors are not overfitting artifacts but generalized logic, evidenced by their robust zero-shot transfer to the visually distinct and semantically rich environments of MapBench and ReasonMap, as well as their consistent effectiveness across model scales.

Ultimately, our findings suggest that while language-centric tasks may benefit significantly from efficiency optimization, visual-spatial reasoning requires the fundamental boundary expansion that verifiable reinforcement learning is particularly well-suited to provide. Furthermore, while RLVR extends this reasoning horizon, we observe that the precise boundary established in synthetic environments does not directly transfer to real-world tasks, where success depends on a more complex interplay of visual semantics and logical depth.

\clearpage

\section*{Limitations}
While our results offer strong evidence for reasoning boundary extension, several limitations remain.
First, our primary analysis relies on the Ariadne maze framework as a proxy for spatial intelligence. While this enables precise boundary verification, future research should incorporate a broader range of real-world scenarios with similarly rigorous definitions of difficulty levels to fully validate the transferability of these boundaries.

Second, due to computational resource constraints, our experiments primarily focus on smaller-scale VLMs from the Qwen2.5-VL and Qwen3-VL series.
We leave it to future work to extend this analysis to a broader range of model variants and larger scales, in order to assess whether frontier models possess sufficient capacity to render these boundaries permeable.

Finally, while we demonstrate effective zero-shot transfer, the induced policy eventually encounters its own complexity horizon in extreme out-of-distribution settings. Moreover, although the weights of many open VLMs are accessible, their pre-training data distributions remain opaque, making it difficult to definitively map the absolute reasoning boundary of existing models beyond empirical probing.

\bibliography{custom}

\clearpage
\appendix
\section{Appendix}
\label{sec:appendix}

\subsection{LLM Usage Statement}

Large language models (LLMs), such as ChatGPT, were used as general-purpose assistive tools during the preparation of this paper. Specifically, LLMs were employed for language refinement and improving the clarity of the manuscript. No part of the research ideation, experimental design, or core scientific contributions relied on LLMs. All scientific content, results, and conclusions were generated and verified by the authors. The authors take full responsibility for the content of this paper, including any text generated with the assistance of LLMs.

\subsection{System Prompt}
\label{sec:prompt}

The following prompt defines a navigation assistant designed for visual path-finding in AlphaMaze. Given a maze image with a green starting cell (``O'') and a red target cell (``T''), the assistant must infer a valid path that passes exclusively through open cells while avoiding black walls.

\begin{tcolorbox}[
    enhanced,              
    colback=gray!5!white,  
    colframe=gray!40!black,
    title={System Prompt for AlphaMaze},
    fonttitle=\bfseries\sffamily,
    fontupper=\small\sffamily,
    boxrule=0.8pt,
    arc=2pt,
    left=6pt, right=6pt, top=6pt, bottom=6pt,
    drop shadow            
]

You are a navigation assistant to solve visual path-finding tasks.

Your goal is to infer a valid path from a visually marked starting point (green cell labeled `O') to a visually marked target (red cell labeled `T') by analyzing the maze image.

\textbf{Rules:}
\begin{itemize}[leftmargin=*, nosep, itemsep=3pt] 
    \item The maze is composed of open paths and impassable black walls.
    \item Movement is only allowed through open paths, not through walls.
    \item You can move one step at a time in the four cardinal directions: \texttt{<|up|>}, \texttt{<|down|>}, \texttt{<|left|>}, \texttt{<|right|>}.
\end{itemize}

\vspace{0.5em}
\textbf{Output Format:}

Think through each step inside \texttt{<think>} and \texttt{</think>} tags.

At each step:
\begin{enumerate}[leftmargin=*, nosep, itemsep=3pt]
    \item Describe your current position based on visual layout and structure (e.g., ``in a corridor'', ``facing a wall'', ``at a crossroad'', ``turning a corner'').
    \item Decide the next move, and explain your reasoning.
    \item Move and continue the path.
\end{enumerate}

After your full reasoning, output only the final movement sequence using the allowed tokens:

\vspace{0.5em}
\centering
\texttt{<|up|>}\texttt{<|down|>}\texttt{<|left|>}\texttt{<|right|>}

\end{tcolorbox}

\subsection{Additional Experimental Results}

\begin{figure*}[t]
\centering
\includegraphics[width=\textwidth]{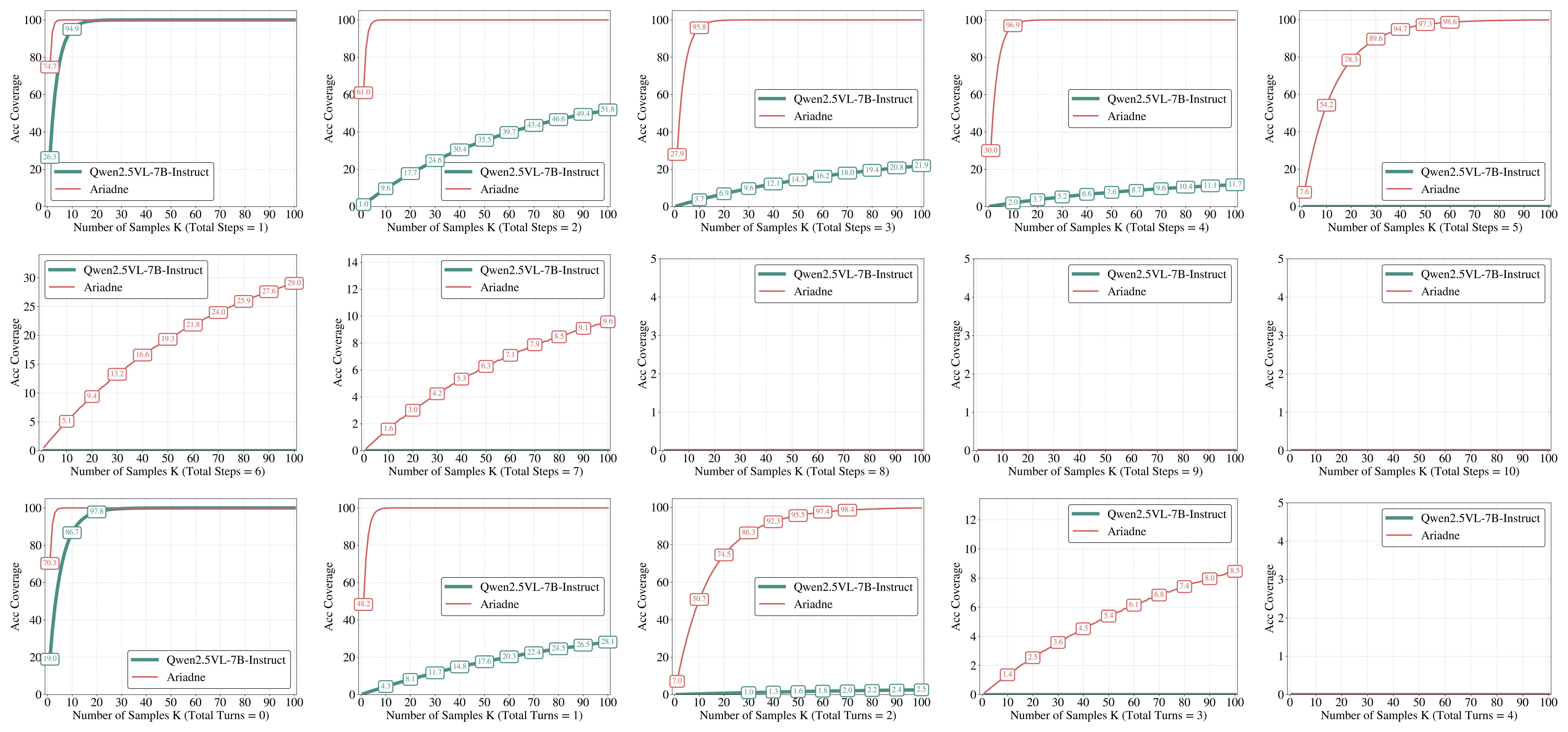}
\caption{
Fine-grained pass@$k$ coverage analysis across step and turn complexity ($k \le 100$, stride = 10).
Coverage is shown as a function of sampling budget for both the base model and Ariadne.
}
\label{fig:acc_turns_steps_fine}
\end{figure*}

To provide a more detailed view of pass@$k$ scaling, we report finer-grained evaluations with denser sampling strides, including a tabulated version in Table~\ref{tab:passk_fine} (focused on low-to-mid budgets) and a continuous visualization in Figure~\ref{fig:acc_turns_steps_fine} (stride = 10 up to pass@100).
Compared to the main text, which uses sparsely spaced evaluation budgets, these results resolve intermediate scaling behavior more precisely.

Across both the table and the figure, the observed trends remain consistent under finer resolution.
In particular, the transition from low to high coverage unfolds smoothly as $k$ increases, and the relative behavior between the base model and Ariadne is preserved across all intermediate budgets.
This consistency indicates that the scaling patterns reported in the main text are not artifacts of specific evaluation points, but reflect stable properties of the underlying policies.

\begin{table*}[t]
\centering
\caption{Fine-grained pass@$k$ coverage across step complexity ($k \le 40$).
Rows denote step counts, and columns report coverage for the base model and Ariadne at different sampling budgets.}
\label{tab:passk_fine}
\resizebox{\textwidth}{!}{
\begin{tabular}{c|cc|cc|cc|cc|cc|cc|cc|cc}
\toprule
\multirow{2}{*}{\textbf{N}} 
& \multicolumn{2}{c|}{pass@1}
& \multicolumn{2}{c|}{pass@5}
& \multicolumn{2}{c|}{pass@15}
& \multicolumn{2}{c|}{pass@20}
& \multicolumn{2}{c|}{pass@25}
& \multicolumn{2}{c|}{pass@30}
& \multicolumn{2}{c|}{pass@35}
& \multicolumn{2}{c}{pass@40} \\
\cline{2-17}
& Base & Ariadne
& Base & Ariadne
& Base & Ariadne
& Base & Ariadne
& Base & Ariadne
& Base & Ariadne
& Base & Ariadne
& Base & Ariadne \\
\midrule
1  & 26.3 & 74.8 & 77.8 & 99.9 & 98.8 & 100.0 & 99.7 & 100.0 & 99.9 & 100.0 & 100.0 & 100.0 & 100.0 & 100.0 & 100.0 & 100.0 \\
2  & 1.1  & 61.1 & 5.0  & 99.0 & 13.8 & 100.0 & 17.7 & 100.0 & 21.3 & 100.0 & 24.6 & 100.0 & 27.6 & 100.0 & 30.4 & 100.0 \\
3  & 0.4  & 27.9 & 1.9  & 80.2 & 5.4  & 99.0  & 6.9  & 99.8  & 8.4  & 99.9  & 9.7  & 100.0 & 10.9 & 100.0 & 12.1 & 100.0 \\
4  & 0.2  & 29.9 & 1.0  & 82.7 & 2.9  & 99.4  & 3.7  & 99.9  & 4.5  & 100.0 & 5.2  & 100.0 & 5.9  & 100.0 & 6.5  & 100.0 \\
5  & 0.0  & 7.6  & 0.0  & 32.6 & 0.0  & 68.6  & 0.0  & 78.4  & 0.0  & 85.0  & 0.0  & 89.4  & 0.0  & 92.6  & 0.0  & 94.8 \\
6  & 0.0  & 0.5  & 0.0  & 2.6  & 0.0  & 7.3   & 0.0  & 9.5   & 0.0  & 11.4  & 0.0  & 13.2  & 0.0  & 14.9  & 0.0  & 16.4 \\
7  & 0.0  & 0.2  & 0.0  & 0.8  & 0.0  & 2.3   & 0.0  & 3.0   & 0.0  & 3.6   & 0.0  & 4.2   & 0.0  & 4.8   & 0.0  & 5.3 \\
8  & 0.0  & 0.0  & 0.0  & 0.0  & 0.0  & 0.0   & 0.0  & 0.0   & 0.0  & 0.0   & 0.0  & 0.0   & 0.0  & 0.0   & 0.0  & 0.0 \\
9  & 0.0  & 0.0  & 0.0  & 0.0  & 0.0  & 0.0   & 0.0  & 0.0   & 0.0  & 0.0   & 0.0  & 0.0   & 0.0  & 0.0   & 0.0  & 0.0 \\
10 & 0.0  & 0.0  & 0.0  & 0.0  & 0.0  & 0.0   & 0.0  & 0.0   & 0.0  & 0.0   & 0.0  & 0.0   & 0.0  & 0.0   & 0.0  & 0.0 \\
\bottomrule
\end{tabular}}
\end{table*}

As shown in Figure~\ref{fig:acc_turns_steps_fine}, the observed trends remain consistent under finer resolution.
In particular, the transition from low to high coverage unfolds smoothly as $k$ increases, and the relative behavior between the base model and Ariadne is preserved across all intermediate budgets.
This suggests that the scaling patterns reported in the main text are not artifacts of specific evaluation points, but reflect stable properties of the underlying policies.

\subsection{Examples from Spatial Reasoning Benchmarks}

We evaluate the generalization of the learned spatial reasoning behaviors on external benchmarks that require navigation and route planning under diverse visual layouts. These benchmarks differ significantly from the synthetic maze environment in both structure and appearance, providing a realistic testbed for assessing whether the learned spatial reasoning strategies transfer beyond the training distribution.

\begin{figure*}[t]
\centering
\includegraphics[width=0.9\linewidth]{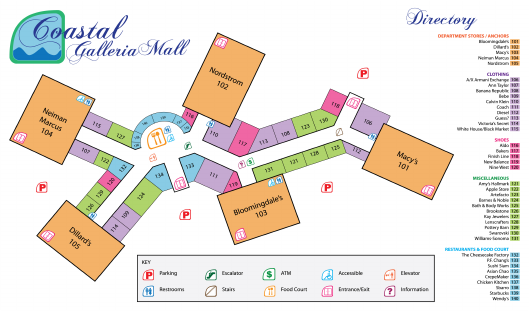}
\caption{
Representative examples from \textbf{MapBench}, which evaluates instruction-following navigation on street maps.
}
\label{fig:mapbench_examples}
\end{figure*}

\begin{figure*}[t]
\centering
\includegraphics[width=0.9\linewidth]{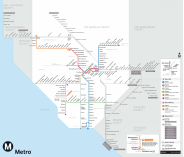}
\caption{
Representative examples from \textbf{ReasonMap}, which assesses fine-grained route planning on transit schematics.
}
\label{fig:reasonmap_examples}
\end{figure*}

Specifically, we consider two representative benchmarks: 1) MapBench, which focuses on instruction-following navigation on street maps, and 2) ReasonMap, which evaluates fine-grained route planning on transit schematics. Figure~\ref{fig:mapbench_examples} and Figure~\ref{fig:reasonmap_examples} illustrate representative examples from these datasets.

\end{document}